%% file: acl_latex.tex
\pdfoutput=1
\documentclass[11pt]{article}

\usepackage{acl}

\usepackage{times}
\usepackage{latexsym}

\usepackage[T1]{fontenc}
\usepackage[utf8]{inputenc}

\usepackage{microtype}

\usepackage{amsmath}
\usepackage{amsfonts}

\usepackage{graphics}
\usepackage{booktabs}

\usepackage{graphicx}
\usepackage{subfig}
\usepackage{caption}

\newcommand{\ra}[1]{\renewcommand{\arraystretch}{#1}}

\newcommand{\clicotea}{\texttt{CLiCoTEA} }

\title{Stop Pre-Training: Adapt Visual-Language Models to Unseen Languages}

\author{Yasmine Karoui\textsuperscript{$\mu$}\thanks{~~Yasmine performed this work while interning at EPFL.} 
\qquad R\'emi Lebret\textsuperscript{$\lambda$}  \qquad Negar Foroutan\textsuperscript{$\lambda$} \qquad Karl Aberer\textsuperscript{$\lambda$} \\
\textsuperscript{$\mu$}Technical University of Munich, Germany \\
\textsuperscript{$\lambda$}EPFL, Switzerland
}
\begin{document}
\maketitle
\begin{abstract}
Vision-Language Pre-training~(VLP) has advanced the performance of many vision-language
tasks, such as image-text retrieval, visual entailment, and visual reasoning.
The pre-training mostly utilizes lexical databases and image queries in English. Previous work  has  demonstrated that the pre-training in English does not transfer well to other languages in a zero-shot setting. However, multilingual pre-trained  language models~(MPLM) have excelled at a variety of single-modal language tasks. In this paper, we propose a simple yet efficient approach to adapt VLP to unseen languages using MPLM.
We utilize a cross-lingual contextualized token embeddings alignment approach to train text encoders for non-English languages. 
Our approach does not require image input and primarily uses machine translation, eliminating the need for target language data.
Our evaluation across three distinct tasks (image-text retrieval, visual entailment, and natural language visual reasoning) demonstrates that this approach outperforms the state-of-the-art multilingual vision-language models  without requiring large parallel corpora. Our code is available at \href{https://github.com/Yasminekaroui/CliCoTea}{https://github.com/Yasminekaroui/CliCoTea}.

% \footnote{Our code is available at \href{https://github.com/Yasminekaroui/CliCoTea}{https://github.com/Yasminekaroui/CliCoTea}}

% This method is able to efficiently train a new textual encoder with relatively low computational cost compared to multilingual VLP, while outperforming the state-of-the-art multilingual vision-language models on image-text retrieval, visual entailment, and natural language visual reasoning.
% rely on image input and depends mainly on machine translation which removes the need for data in the target language.
% We find that

\end{abstract}

\input{intro.tex}
\input{method.tex}
\input{experiments.tex}
\input{conclusion.tex}

\input{limitations.tex}

\bibliography{anthology,custom}
\bibliographystyle{acl_natbib}

\appendix

%\section{Example Appendix}
\input{appendix.tex}

\label{sec:appendix}

\end{document}

%% file: intro.tex
\section{Introduction}

% \subsection{Context}
Inspired by the recent advancements in language model pre-training, Vision-Language Pre-trained Models~(VLPMs) have demonstrated state-of-the-art performance across a wide range of vision-language~(VL) tasks such as text-to-image retrieval, visual reasoning, visual entailment, and visual QA~\cite{chen2020uniter, li2021align, li2022blip}.
\input{model_overview.tex}
% The multilingual and multimodal domains have recently benefited from the enormous success of language model pre-training. A wide spectrum of users, including non-English speakers, have benefited from cutting-edge language technology thanks to modern multilingual pre-training approaches. The state-of-the-art monolingual pre-trained language models, like BERT~\cite{devlin2018bert} or RoBERTa~\cite{liu2019roberta} have now their multilingual expansion: mBERT~\footnote{\url{https://github.com/google-research/bert/blob/master/multilingual.md}} and  XLM/XLM-R~\cite{conneau2019unsupervised}, respectively.
% Similar to this, multimodal pre-training expands the range of tasks for which pre-trained models can be used. Some powerful pre-trained  multimodal models recently came out such as UNITER \cite{chen2020uniter}, ALBEF \cite{li2021align} and BLIP \cite{li2022blip}.

However, extending VLPMs to multilingual scenarios is still challenging. On one hand, the majority of these models are trained on monolingual~(English) corpora and thus cannot perform well for other languages. 
On the other hand, the multilingual pre-trained language models~\cite{devlin2018bert, conneau2019unsupervised} cannot handle vision data~(e.g., images or videos) directly.

% Since these two expansions have proven their effectiveness in solving both multilingual and multimodal problems, the next step was to combine both directions.
% But, extending these models to multilingual multimodal use cases is still problematic. First because multilingual pre-trained models are not able to deal with visual input, and second the majority of the multimodal models are pre-trained on English datasets hence cannot perform well on non-English languages.
% Thus, large high quality multilingual multimodal data is a necessity to combine multilingual pre-training and multimodal pre-training. Nevertheless, there are just a few multilingual multimodal datasets, and they cover a small amount of languages. On the contrary, relatively large datasets are available for multilingual pre-training (e,g. Wikipedia in 100 languages) and multimodal pre-training. Furthermore, constructing large multilingual dataset from English multimodal corpora by using machine translation engines is time and computationally expensive.
Lately, there have been attempts (M$^3$P, nUNITER, UC$^2$) to pivot on images or English texts to align multilingual representations with vision features~\cite{chen2020uniter, ni2021m3p, zhou2021uc2}.
However, a recent benchmark on multilingual multimodal pre-training~(IGLUE)~\cite{bugliarello2022iglue} shows that although these models achieve promising zero-shot cross-lingual transfer performance on some VL tasks, they still fall short in comparison to the ``translate-test'' baseline (using an English-only VLPM on the translations of the text examples).

% Two pioneering works, M3P \cite{ni2021m3p} and UC2 \cite{zhou2021uc2}, propose to pivot on images or  English texts to align multilingual representations with vision features. However, IGLUE \cite{bugliarello2022iglue} which is a recent benchmark on multilingual multimodal pre-training  reveals the limitations of these models: while performing well on zero-shot cross-lingual transfer performance on some multilingual multimodal tasks, they underperform "translation test" evaluation which is a simple baseline which translates multilingual multimodal test set into English and adopt a powerful English vision-language model for inference. In contrast, some multilingual pre-trained models e.g. XLM-R \cite{conneau2019unsupervised} achieve better results than the "translation test" evaluation in a lot of languages and in a wide range of applications.
A more recent work~(CCLM) achieves promising performance on the IGLUE benchmark by exploiting massive parallel text and image-text corpora to pre-train a VL model~\cite{zeng2022cross}.
This approach is motivated by a key observation that multilingual and multimodal pre-training essentially achieves the same goal of aligning two different views of the same object into a common semantic space.
Although this framework performs well on the IGLUE benchmark, it requires a large amount of parallel data.
Its pre-training phase relies on 19M multilingual parallel sentence pairs extracted from WikiMatrix~\cite{schwenk-etal-2021-wikimatrix}, jointly trained with 4 million image-text pairs in multiple languages.
% Most recent work  \cite{zeng2022cross}  pre-train a Cross-lingual Cross-modal Language Model (CCLM) with the cross-view language modeling framework. Results on IGLUE show that CCLM significantly outperforms the prior state-of-the-art with an average improvement of ~10\%. Moreover, CCLM is the first multilingual multimodal model that outperforms the "translation test" baseline by zero-shot cross-lingual transfer.
% However, although this framework has proven its efficiency on IGLUE benchmark, it is computationally expensive. In fact, the pre-training relies on 19M multilingual parallel sentence pairs extracted from WikiMatrix \cite{schwenk-etal-2021-wikimatrix}.
% \newline\vspace{5mm}
% \subsection{Problem Definition}\label{chapter:Problem_def}

In this work, we are proposing a simple yet efficient way to adapt VLP models to unseen languages without requiring large parallel corpora.
We propose to align a VLPM monolingual text encoder (achieving start-of-the-art performance on English downstream VL tasks) with a multilingual pre-trained language model~(e.g., mBERT), using only small in-domain parallel text corpus.
The recent progress in Neural Machine Translation~(NMT) has enabled us to create such a parallel corpus from automatically translating the data from English to any other language, even for low-resource languages~(i.e., Swahili).
However, since our approach relies on token alignment, it is robust to errors made by NMT.
Our zero-shot evaluation across three of the four IGLUE tasks shows that the proposed method achieves state-of-the-art results while using small set of in-domain parallel sentences. The key steps of our approach are illustrated in Figure~\ref{fig:method_visualization}.

% Although several Vision-Language Pre-trained Models (VL-PTMs) have been introduced inspired by BERT architecture, these VL-PTMs are trained on large image-text datasets that are generally in English. On the other hand, cross-lingual transfer learning focuses only on text. 

% In this paper, we find an approach to quickly and efficiently adapt vision-language pre-training to unseen languages, especially low-resource languages. For this aim, we consider three tasks from the IGLUE benchmark datasets for evaluating our method in zero-shot cross-lingual transfer setting.

%% file: model_overview.tex
\begin{figure}[!ht]
\centering
\subfloat[Cross-Lingual Text Alignment]{
        \label{fig:methodology_step1}
        {\includegraphics[width=1\columnwidth]{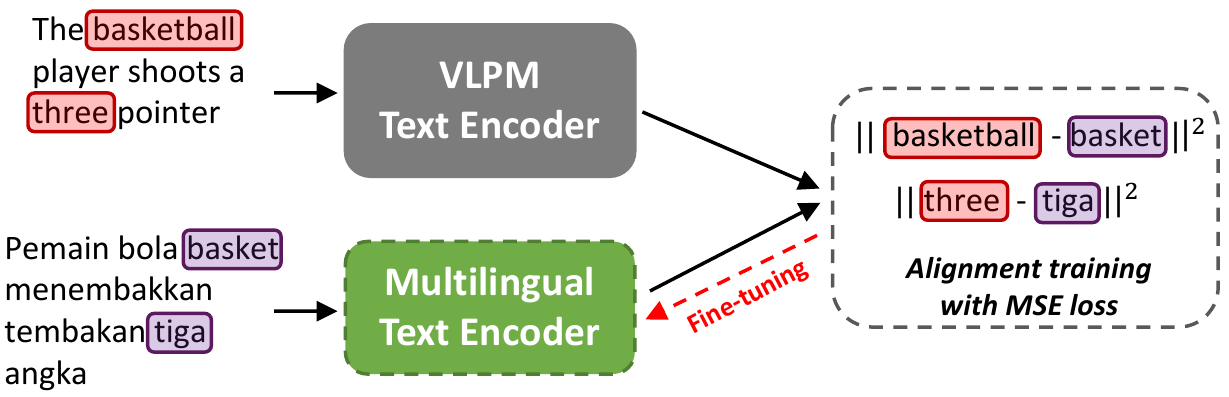}}
    }

% \vspace{-2mm}

\subfloat[Zero-shot Transfer on VL downstream task]{
        \label{fig:methodology_step2}
        {\includegraphics[width=1\columnwidth]{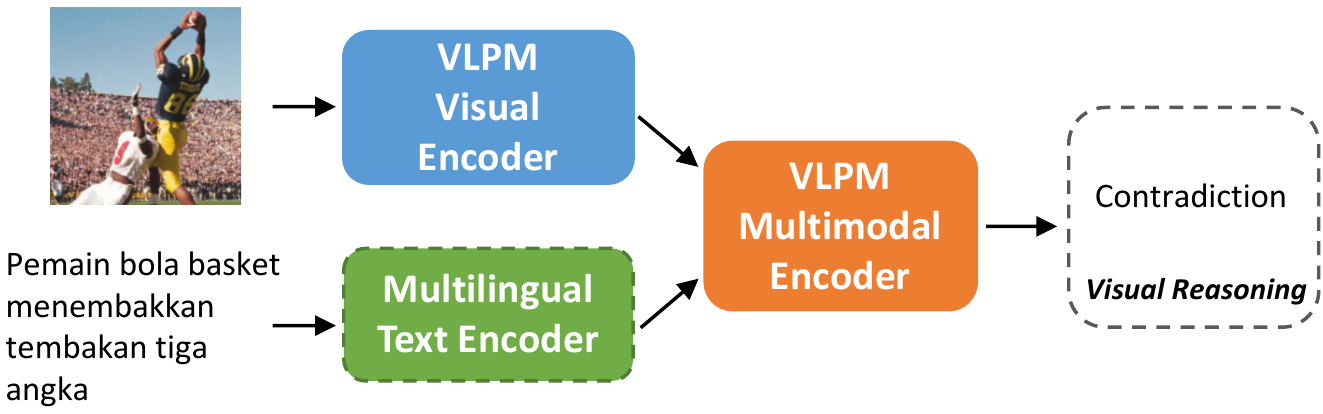}}
    }
\caption{Overview of our approach. 
We adapt the text encoder of a monolingual VL model to an unseen language (a). Then we use the adapted model for a VL downstream task in a zero-shot setting (b). 
}
\label{fig:method_visualization}
\vspace{-6mm}
\end{figure}

%% file: method.tex
\section{\clicotea: Cross-Lingual Contextualised Token Embedding Alignment}
\label{sec:method}

We propose \clicotea, an approach to transfer a monolingual vision-language~(VL) pre-trained model in one language $L_1$ where there is an abundant number of training pairs of image and text (i.e., English) to a second language $L_2$. As we focus in this paper on the zero-shot setting, we do the transfer after fine-tuning the pre-trained monolingual VL model on a downstream task $t$, where training samples are available in language $L_1$. 

\clicotea consists of six steps:
\begin{enumerate}
    \item\label{pipeline:pretrain} Pre-train a monolingual VL model on a massive collection of image-text pairs, where text is written in language $L_1$.
    \item\label{pipeline:finetune} Fine-tune the VL pre-trained model on the downstream task $t$ in language $L1$.
    \item\label{pipeline:corpus} Create a parallel text corpus by translating the training set from step~\ref{pipeline:finetune} in the target language $L_2$. Note that this step can be done automatically using neural machine translation.
    \item\label{pipeline:token} Create a list of aligned tokens for each (potentially noisy) parallel sentence using a token alignment model.
    \item\label{pipeline:transfer} Cross-lingual transfer by aligning contextualised token embeddings. As illustrated in Figure~\ref{fig:method_visualization}a, it transfers the VL fine-tuned model to the new language $L_2$ by aligning a pre-trained multilingual LM (e.g., mBERT or XLM-R) with the text encoder of the VL pre-trained model using the list of aligned tokens created in step~\ref{pipeline:token}.
    \item\label{pipeline:zeroshot} Zero-shot transfer to $L_2$ by swapping the monolingual text encoder from the VL pre-trained model with the aligned multilingual text encoder learned in step~\ref{pipeline:transfer}. An example of visual reasoning in Indonesian is illustrated in Figure~\ref{fig:method_visualization}b.
\end{enumerate} 

In practice, steps \ref{pipeline:pretrain} and \ref{pipeline:finetune}
are the most computationally expensive. Therefore, we propose to adapt VL fine-tuned models to new languages by only doing the steps from \ref{pipeline:corpus} to \ref{pipeline:transfer} which can be computed in a few hours on a single GPU.

We note that \clicotea could be used with any multimodal pre-trained model where one of the modalities is a monolingual text encoder. We focus in this paper on VL models, but \clicotea could be applied for instance to a language-knowledge model such as GreaseLM~\cite{zhang2021greaselm} or DRAGON~\cite{yasunaga2022deep}.  

%% file: experiments.tex
\section{Experiment}

\subsection{Pre-trained Models} 

\paragraph{Vision-Language Model} 
In step~\ref{pipeline:pretrain} of \clicotea, we use the Align BEfore Fuse (ALBEF) framework\footnote{Code and models are available at \url{https://github.com/salesforce/ALBEF}.}~\cite{li2021align} as our Vision-Language Pre-trained Model~(VLPM). ALBEF has been fine-tuned on multiple downstream VL tasks and achieves state-of-the-art performance. We use the ALBEF fine-tuned models in step~\ref{pipeline:finetune} for the downstream tasks described in Section~\ref{sec:tasks}. Unlike other competitive VL pre-trained models (such as BLIP~\cite{li2022blip}) that inject visual information by inserting cross-attention for each transformer block of the text encoder, ALBEF first encodes the image and text independently with a detector-free image encoder and a text encoder. Then it uses a multimodal encoder to fuse the image features with the text features through cross-modal attention. All encoders are based on transformer networks with the text encoder being a 6-layer transformer initialised using the first 6 layers of the BERT\textsubscript{base}.
We thus extract this 6-layer text encoder for cross-lingual transfer training in step~\ref{pipeline:transfer}.  

\paragraph{Multilingual Language Model} 
As a multilingual pre-trained language model, we use the multilingual BERT~(mBERT)\footnote{Available on HuggingFace hub at~\url{https://huggingface.co/bert-base-multilingual-cased}.}~\cite{devlin2018bert}. It has been trained on the top 104 languages with the largest Wikipedia using a masked language modeling~(MLM) objective and has demonstrated remarkable zero-shot cross-lingual transfer capabilities~\cite{wu2019beto, pires2019multilingual, hu2020xtreme, conneau2018xnli}. We extract the first 6-layer transformer to be aligned with the text encoder of ALBEF in step~\ref{pipeline:transfer}.

\subsection{Implementation Details}

\paragraph{Word Alignment}
\label{sec:word_align}
Since the  parallel sentences do not contain word-level alignment information, in step~\ref{pipeline:token} of \clicotea we utilize \texttt{awesome-align}\footnote{\url{https://github.com/neulab/awesome-align}}~\cite{dou2021word} which is a tool that automatically extracts word alignments from mBERT. The generated word pairs are then filtered for keeping only one-to-one, one-to-many or many-to-one alignments and removing many-to-many alignments. This is done for all languages except Chinese because otherwise less than 3\% of the training data would remain in the set. The advantage of this filtering is twofold: a) it removes the noise from the matching word pairs; b) it reduces the training time and computation. For words that are split into sub-word tokens, we consider either the left-most token embedding alignment (i.e., the first sub-word token of a word) or, the average embedding across all sub-word tokens.

\paragraph{Contextualised Token Alignment Training} 
Given a set of aligned contextual word pairs extracted from parallel sentences, we define $\{\mathrm{x}_i, \mathrm{y}_i\}^n_{i=1}$, where $\mathrm{x}_i \in \mathbb{R}^{d}$ is the contextualised embedding of token $i$ in the target language (obtained from mBERT), and $\mathrm{y}_i \in \mathbb{R}^{d}$ is the contextualised embedding of its alignment in the source language (obtained from the fine-tuned ALBEF)\footnote{Note that the special $\mathtt{[CLS]}$ token is always included.}.
In step~\ref{pipeline:transfer} of \clicotea, we minimise the following training objective: $\sum_{i=1}^n || \mathrm{x}_i - \mathrm{y}_i||^2$.

% \begin{align*}
%     \sum_{i=1}^n || \mathrm{x}_i - \mathrm{y}_i||^2.
% \end{align*}
The parameters of the source language encoder are frozen, while the ones of the target language encoder are fine-tuned at training time. The learning rate is set to $5.10^{-5}$. The batch size is set to 128. These hyperparameters are set through the NLVR2, Flickr30k, SNLI validation sets, for each task respectively. For each target language, the training is done on a single GeForce GTX TITAN X in a few hours.

\paragraph{Data Augmentation} 
\label{sec:data_aug}
As multilingual language models are generally pre-trained on the source language $L_1$, the contextualised token alignment can be trained not only with sentences from the target language $L_2$, but also with sentences from the source language $L_1$. This strategy doubles the training size, and consequently, the training time but it could be used with tasks where the number of available training sentences is limited.

\subsection{Downstream Tasks}
\label{sec:tasks}

In step~\ref{pipeline:zeroshot}, we evaluate \clicotea on three tasks from the IGLUE benchmark\footnote{We do not include the Cross-lingual Grounded Question Answering task (xGQA~\cite{pfeiffer2021xgqa}) as it requires aligning the answer decoder too. We leave it as future work.} in the zero-shot setting:
\begin{itemize}
    \item \textbf{xFlickr\&CO}: The dataset is composed of 1000 images from Flickr30K~\cite{plummer2015flickr30k} and 1000 images from MSCOCO dataset~\cite{lin2014microsoft}. These images come along with croudsourced image captions in 6 different languages. xFlickr\&CO is a \emph{retrieval} task dataset. It is composed of two subtasks: image-to-text retrieval (TR) and text-to-image retrieval~(IR).
    \item  \textbf{XVNLI}: The dataset consists in merging SNLI hypothesis with Flickr30K \cite{plummer2015flickr30k} images and translate the test set in four languages.
    The task is called \emph{visual entailment} (VE) which is a fine-grained reasoning task to determine whether a text hypothesis ``contradicts'', ``entails'', or is ``neutral'' with respect to an image.
    \item  \textbf{MaRVL}: The dataset is a multilingual expansion of NLVR2 dataset \cite{suhr2017corpus}, with images related to concepts of five languages and cultures. The task is called \emph{visual reasoning}~(VR) which consists in determining whether a statement is correct given a pair of images.
\end{itemize}

\begin{table}[h]
\ra{1.3}
\resizebox{\columnwidth}{!}{%
\centering
\begin{tabular}{@{}lccc@{}}
\toprule
\multicolumn{1}{l}{\textbf{Step}}&\textbf{Retrieval} & \textbf{VE} & \textbf{VR} \\
\midrule
\textit{Fine-tuning} & Flickr30K &  SNLI &  NLVR2\\ 
\textit{Alignment} & Flickr30K$^*$  & SNLI$^*$ & NLVR2$^*$  \\
\textit{Zero-shot Test} &  xFlickr\&CO & XVNLI & MaRVL \\
\bottomrule
\end{tabular}%
}
\caption{The datasets used in the different steps of \clicotea. Translated train and validation captions are denoted with $*$.}

\label{tab:datasets}
\end{table}

Table~\ref{tab:datasets} shows the datasets used for a) fine-tuning the monolingual VL pre-trained model in step~\ref{pipeline:finetune}, b) training the alignment of contextualised token embeddings in step~\ref{pipeline:transfer}, and c) testing the zero-shot cross-lingual transfer in step~\ref{pipeline:zeroshot}.
For creating the parallel corpus in step~\ref{pipeline:corpus}, all datasets used for fine-tuning the monolingual pre-trained VL model are translated to the corresponding test dataset languages from the IGLUE benchmark using GoogleTrans Python API\footnote{\url{https://pypi.org/project/googletrans/}}. Statistics about the translation datasets can be found in Section~\ref{sec:align_dataset}. MaRVL being the smallest dataset, the data augmentation strategy described in Section~\ref{sec:data_aug} is applied only for this task. Detailed results on data augmentation can be found in Section~\ref{sec:data_aug}.

\subsection{Experimental Results}
\begin{table}[h!]
\ra{1.3}
\centering
\resizebox{\columnwidth}{!}{%
\begin{tabular}{@{}lrcrcrr@{}}
\toprule
\textbf{Model} & \multicolumn{1}{c}{\textbf{VE}} &&\multicolumn{1}{c}{\textbf{VR}} && \multicolumn{2}{c}{\textbf{Retrieval}}\\
 \cmidrule{2-2} \cmidrule{4-4} \cmidrule{6-7}
& \multicolumn{1}{c}{XVNLI} & & \multicolumn{1}{c}{MaRVL} && \multicolumn{2}{c}{xFlickr\&CO} \\ 
& & & & & \multicolumn{1}{c}{IR} & \multicolumn{1}{c}{TR} \\
\midrule
mUNITER & 53.69 && 53.72 && 8.06 & 8.86\\
xUNITER & 58.48 && 54.59 && 14.04 & 13.51\\
UC$^{2}$ & 62.05 && 57.28 && 20.31 & 17.89 \\ 
M$^{3}$P & 58.25 && 56.00 && 12.91 & 11.90 \\
CCLM\textsubscript{3M} & 74.64 && 65.91 && 67.35 & 65.37\\
CCLM\textsubscript{4M} & 73.32 && 67.17 && \textbf{76.56} & \textbf{73.46}\\
\midrule
\clicotea & \textbf{78.15} && \textbf{68.09} && 67.45 & 65.07  \\
\bottomrule
\end{tabular}%
}
\caption{Zero-shot performance on IGLUE benchmark. Recall@1 and Accuracy are reported for retrieval tasks (xFlickr\&CO) and understanding tasks (XVNLI, MaRVL) respectively. Results of compared models are directly copied from \citet{zeng2022cross}.}
\label{tab:iglue}
\end{table}

Results reported in Table~\ref{tab:iglue} shows that \clicotea outperforms the state-of-the-art CCLM models for all downstream tasks except retrieval. The larger improvement against CCLM models is obtained in visual entailment with an increase of almost 5\%. The superiority of \clicotea is especially high for Spanish (+7.68\%), as can be seen from Table~\ref{tab:xvnli_results} in Section~\ref{sec:res_ve}.
The average performance on visual reasoning is similar to CCLM, but \clicotea significantly outperforms CCLM by $\pm 4\%$ on the low-resource languages such as Tamil and Swahili (results per language can be seen in Table~\ref{tab:marvl_results} in Section~\ref{sec:res_vr}). 
For retrieval, \clicotea outperforms all models except CCLM\textsubscript{4M}. It is worth mentioning that, unlike the other models, CCLM\textsubscript{4M} has been pre-trained on COCO which could explain its superiority on Flickr\&CO dataset. More details about the results on retrieval can be found in Section~\ref{sec:res_retrieval}.

%% file: conclusion.tex
\section{Conclusion}
In this paper, we present \clicotea an approach for adapting Vision-Language pre-trained models to unseen languages. Unlike other approaches that rely on an expensive pre-training phase (both in terms of data and computation), our approach adapts the contextualised token embeddings of a multilingual pre-trained language model by aligning them with the contextualised token embeddings of the VLPM text encoder.
By aligning ALBEF text encoder with mBERT, we show that \clicotea outperforms CCLM, which exploits massive parallel text and image-text corpora. \clicotea achieves start-of-the-art performance on visual entailment and visual reasoning, with an increase of almost 5\% on visual entailment. It also demonstrates its effectiveness, especially for low-resource languages, as it does not require large corpora to do the adaptation. 
% This work also shows that machine translation is a potential solution to solve the language bias problem. 

%% file: limitations.tex
\section{Limitations}
% For our experiments, we use mBERT as a multilingual pre-trained language model. 
The general performance of \clicotea could be improved with a better MPLM than mBERT, such as XLM-R which has a larger token vocabulary and has been pre-trained on a much larger dataset.
Our approach is currently not applicable to generation tasks where a multilingual text decoder is needed to generate text in unseen languages. We leave this adaptation for future work.
Unlike the statement made in \citet{zeng2022cross}, current multilingual VL models still do not surpass the \emph{Translate-Test} baseline of the tasks from IGLUE benchmark. The performance of \clicotea is promising but the best scores are still obtained when translating everything to English and using the (English-only) ALBEF model. The smallest difference in accuracy on MaRVL dataset between \clicotea and ALBEF with \emph{Translate-Test} is obtained in Swahili (-2\%), while the gap is much larger (around -6\%) for the other languages.
Outperforming the \emph{Translate-Test} achieved by ALBEF still remains an open challenge, especially for high-resource languages.

% Another limitation is 

% \begin{itemize}
%     \item Our approach rely on machine translations that could be noisy and help propagate error.
%     \item Our method works only for non generation tasks where no decoder is needed.
% \end{itemize}

%% file: appendix.tex
\section{Appendix}

\subsection{Details of Alignment Datasets}
\label{sec:align_dataset}

Tables~\ref{tab:n_tokens_flickr}, \ref{tab:n_tokens_snli}, and \ref{tab:n_tokens_marvl} show the average number of aligned tokens extracted from the translated sentences of Flickr30k, SNLI, and NLVR2, respectively. 

\begin{table}[h]
    \centering
    \begin{tabular}{@{}lrr@{}}
    \toprule
        \multicolumn{1}{c}{Language} & \multicolumn{1}{c}{Total number }& \multicolumn{1}{c}{Avg. number of} \\ 
        &\multicolumn{1}{c}{of sentences}& \multicolumn{1}{c}{aligned tokens}\\
        \midrule
        German & 144935 & 8.74\\
        Spanish & 144990 & 10.04\\
        Indonesian & 144858 & 7.46\\
        Russian & 144526 & 6.44\\
        Turkish & 143664 & 4.83\\
        \bottomrule
    \end{tabular}
    \caption{Statistics about Flickr30k translation set.}
    \label{tab:n_tokens_flickr}
\end{table}

\begin{table}[h]
    \centering
    \begin{tabular}{@{}lrr@{}}
    \toprule
        \multicolumn{1}{c}{Language} & \multicolumn{1}{c}{Total number }& \multicolumn{1}{c}{Avg. number of} \\ 
        &\multicolumn{1}{c}{of sentences}& \multicolumn{1}{c}{aligned tokens}\\
        \midrule
        Arabic & 513683 & 2.95\\
        Spanish & 549785 & 6.31\\
        French & 549260 & 5.78\\
        Russian & 524308 & 3.60\\
        \bottomrule
    \end{tabular}
    \caption{Statistics about SNLI translation set.}
    \label{tab:n_tokens_snli}
\end{table}

\begin{table}[h]
    \centering
    \begin{tabular}{@{}lrr@{}}
    \toprule
        \multicolumn{1}{c}{Language} & \multicolumn{1}{c}{Total number }& \multicolumn{1}{c}{Avg. number of} \\ 
        &\multicolumn{1}{c}{of sentences}& \multicolumn{1}{c}{aligned tokens}\\
        \midrule
        Indonesian & 86325 & 8.27\\
        Swahili & 85415 & 5.46\\
        Tamil & 85241 & 4.53\\
        Turkish & 85050 & 5.42\\
        Chinese & 86373 & 10.76\\
        \bottomrule
    \end{tabular}
    \caption{Statistics about NLVR2 translation set.}
    \label{tab:n_tokens_marvl}
\end{table}

\subsection{Results on Retrieval}
\label{sec:res_retrieval}

Zero-shot performance on the Flickr\&CO dataset, the image-text and text-image retrieval tasks from the IGLUE benchmark, for four available languages (DE: German, ES: Spanish, ID: Indonesian, RU: Russian, TR: Turkish) are reported in Table~\ref{tab:flickrco_ir} and Table~\ref{tab:flickrco_tr}, respectively.
\clicotea outperforms all models except CCLM\textsubscript{4M}. Compared with CCLM\textsubscript{3M}, CCLM\textsubscript{4M} has been trained with 1M additional image-text pairs from Visual Genome and COCO datasets. The gap in performance between the two models on retrieval tasks suggests that pre-training with COCO text-image pairs gives a clear advantage to CCLM\textsubscript{4M} as Flickr\&CO contains 1000 images from COCO, while all other models have been fine-tuned only on Flickr30K.

\begin{table}[h!]
\centering
\ra{1.3}
\resizebox{\columnwidth}{!}{%
\begin{tabular}{@{}lccccc@{}}
\toprule
\textbf{Model} & \multicolumn{5}{c}{\textbf{Language}}\\
\cmidrule{2-6}
 &  \multicolumn{1}{c}{DE}& \multicolumn{1}{c}{ES} & \multicolumn{1}{c}{ID} & \multicolumn{1}{c}{RU} & \multicolumn{1}{c}{TR}  \\
\hline 
mUNITER & 12.05 & 13.15 & 5.95 & 5.85 & 1.75\\
xUNITER & 14.55 & 16.10 & 16.50 &  15.90 & 9.05\\
UC$^2$ & 28.60 & 15.95 & 14.60 & 20.00 & 7.15\\
M${^3}$P & 13.35  & 13.40 & 13.20 & 15.95 & 7.75\\
CCLM\textsubscript{3M} & 67.67 & 71.23 & 62.38 & 72.83 &  55.15\\
CCLM\textsubscript{4M} & \textbf{73.65} & \textbf{79.62} & \textbf{69.50} & \textbf{80.65} & \textbf{65.08}
\\
\midrule
% \clicotea old & 60.10 & 62.85  & 49.82 & 54.53& 49.65\\

\clicotea  & 61.48 & 74.50 & 64.98 & 73.50 & 62.80 \\
% \midrule 
% Translation test &74.38&74.63&
% 84.10&84.10&
% 75.38&76.43&
% 86.75&86.90&
% 80.65& 81.35 & 
% 80.25 & 80.68\\ 
\bottomrule
\end{tabular}
}
\caption{Zero-shot performance on multi-lingual image-text retrieval with Flickr\&CO dataset. Recall@1 is reported.}
\label{tab:flickrco_ir}
\end{table}

\begin{table}[h!]
\centering
\ra{1.3}
\resizebox{\columnwidth}{!}{%
\begin{tabular}{@{}lccccc@{}}
\toprule
\textbf{Model} & \multicolumn{5}{c}{\textbf{Language}}\\
\cmidrule{2-6}
 &  \multicolumn{1}{c}{DE}& \multicolumn{1}{c}{ES} & \multicolumn{1}{c}{ID} & \multicolumn{1}{c}{RU} & \multicolumn{1}{c}{TR}  \\
\hline 
mUNITER & 11.85 & 13.05 & 7.55 & 6.80 & 3.25\\
xUNITER & 13.25 & 15.10 & 16.75 & 14.80 &  10.05\\
UC$^2$ & 23.90 & 15.30 & 13.60 & 16.75 & 6.95\\
M${^3}$P & 11.85 & 12.15 & 12.10 & 14.45 & 8.35 \\
CCLM\textsubscript{3M} & 66.88 & 68.58 & 60.33 & 69.90 & 54.22\\
CCLM\textsubscript{4M} & \textbf{73.60} & \textbf{78.38} & \textbf{67.67} & \textbf{80.35} & \textbf{63.22} \\
\midrule
\clicotea & 70.34 & 71.42 & 57.77 & 69.80 & 56.00 \\
% new & 56.92 & 71.05 &56.53 & 66.2 & 53.4 \\
% \midrule 
% Translation test &74.38&74.63&
% 84.10&84.10&
% 75.38&76.43&
% 86.75&86.90&
% 80.65& 81.35 & 
% 80.25 & 80.68\\ 
\bottomrule
\end{tabular}
}
\caption{Zero-shot performance on multi-lingual text-image retrieval with Flickr\&CO dataset. Recall@1 is reported.}
\label{tab:flickrco_tr}
\end{table}

% \begin{table*}[h!]
% \centering
% \ra{1.3}
% \scalebox{0.95}{
% \begin{tabular}{@{}lcccccccccc@{}}
% \toprule
% \textbf{Model} & \multicolumn{10}{c}{\textbf{Language}}\\
% \cmidrule{2-11}
%  &  \multicolumn{2}{c}{DE}& \multicolumn{2}{c}{ES} & \multicolumn{2}{c}{ID} & \multicolumn{2}{c}{RU} & \multicolumn{2}{c}{TR}  \\
%  & TR&IR& TR&IR& TR&IR& TR&IR& TR&IR\\
% \hline 
% mUNITER & 11.85& 12.05&
% 13.05&13.15&
% 7.55& 5.95&
% 6.80&5.85&
% 3.25&1.75\\
% xUNITER & 13.25&14.55  &
% 15.10&16.10 &
% 16.75&16.50&
% 14.80& 15.90&
% 10.05& 9.05\\
% UC$^2$ &23.90 &28.60 &
% 15.30&15.95 &
% 13.60& 14.60&
% 16.75&20.00&
% 6.95&7.15\\
% M${^3}$P&11.85 &13.35  &
% 12.15&13.40&
% 12.10&13.20&
% 14.45&15.95&
% 8.35&7.75\\
% CCLM_{3M} &66.88 &67.67 &
% 68.58&71.23 &
% 60.33&62.38&
% 69.90& 72.83&
% 54.22& 55.15
% \\
% CCLM_{4M}& \textbf{73.60}&\textbf{73.65}& 
% \textbf{78.38}&\textbf{79.62}  &
% \textbf{67.67}&\textbf{69.50} &
% \textbf{80.35}&\textbf{80.65}&
% \textbf{63.22}& \textbf{65.08}
% \\
% \midrule
% \clicotea old &70.34  & 60.10&
% 71.42 &62.85  &
% 57.77& 49.82 &
% 69.80& 54.53&
% 56.00& 49.65
% \\

% \clicotea  &  & &
%  &74.5  &
% & 64.98 &
% &&
% & 62.8& 
% \\
% % \midrule 
% % Translation test &74.38&74.63&
% % 84.10&84.10&
% % 75.38&76.43&
% % 86.75&86.90&
% % 80.65& 81.35 & 
% % 80.25 & 80.68\\ 
% \bottomrule
% \end{tabular}
% }
% \caption{Zero-shot performance on multi-lingual image-text retrieval with Flickr\&CO dataset. Recall@1 is reported.}
% \label{tab:flickrco_results}
% \end{table*}

\subsection{Results on Natural Language Visual Reasoning}
\label{sec:res_vr}

Table~\ref{tab:marvl_results} shows the zero-shot performance on the MaRVL dataset, and the natural language visual reasoning task from the IGLUE benchmark, for all available languages (ID:~Indonesian, SW:~Swahili, TA:~Tamil, TR:~Turkish, ZH:~Chinese). 

\begin{table}[h!]
\ra{1.3}
\centering
\resizebox{\columnwidth}{!}{%
\begin{tabular}{@{}lrrrrr@{}}
\toprule
\textbf{Model} & \multicolumn{5}{c}{\textbf{Language}}\\
\cmidrule{2-6}
 & \multicolumn{1}{c}{ID} & \multicolumn{1}{c}{SW} & \multicolumn{1}{c}{TA} & \multicolumn{1}{c}{TR}& \multicolumn{1}{c}{ZH}  \\
 \midrule
mUNITER & 54.79 &51.17 & 52.66&  54.66 & 55.34\\
xUNITER & 55.14& 55.51 &53.06 & 56.19& 53.06\\
UC$^2$ & 56.74& 52.62&  60.47 & 56.70& 59.88\\
M${^3}$P& 56.47& 55.69 &56.04 & 56.78& 55.04\\
CCLM\textsubscript{3M} & 67.81& 61.55& 60.28& 69.60& \textbf{70.52} \\
CCLM\textsubscript{4M} & \textbf{71.66} & 67.21 & 60.36 &66.75& 69.86 \\
\midrule
\clicotea & 69.55 & \textbf{71.30} & \textbf{63.93} & \textbf{70.72} &64.93\\
% \midrule
% Translation test &   74.1  & 73.7 &  69.9 & 77.8 & 77.0 & 74.5  \\ 
\bottomrule
\end{tabular}
}
\caption{Zero-shot performance on visual reasoning with MaRVL dataset. Accuracy is reported.}
\label{tab:marvl_results}
\end{table}

As MaRVL is the smallest dataset among the three tasks from IGLUE, we apply the data augmentation for training the alignment as described in Section~\ref{sec:data_aug}. Results reported in Table~\ref{tab:marvl_lang} show that there is drop of 3.35\% for Turkish, and 9.99\% for Chinese when training only using the target language $L_2$, while there is no significant difference for the three other languages (Indonesian, Swahili, and Tamil).
As explained in Section~\ref{sec:word_align}, our noise filtering technique does not work well with Chinese. Aligning the English sentences with half of the original training set helped the model infer knowledge from English and reduced the number of wrong matching words. For Turkish, the increase in performance could be explained by the similarity between the two alphabets.

\begin{table}[h!]
\ra{1.3}
\centering
\resizebox{\columnwidth}{!}{%
\begin{tabular}{@{}lrrrrr@{}}
\toprule
\textbf{Training Set} & \multicolumn{5}{c}{\textbf{Language}}\\
\cmidrule{2-6}
  & \multicolumn{1}{c}{ID} & \multicolumn{1}{c}{SW} & \multicolumn{1}{c}{TA} & \multicolumn{1}{c}{TR}& \multicolumn{1}{c}{ZH}  \\
 \midrule
 $L_1$ & 69.55 & 71.30 & 63.45 & 67.37 & 54.94\\
 $L_1+L_2$ & 68.53 & 70.31 & 63.93 &70.72 & 64.93\\
\bottomrule
\end{tabular}
}
\caption{Zero-shot performance of \clicotea on visual reasoning with MaRVL dataset using monolingual ($L_1$) or bilingual ($L_1+L_2$) alignment training. Accuracy is reported.}
\label{tab:marvl_lang}
\end{table}

\subsection{Results on Visual Entailment}
\label{sec:res_ve}

Zero-shot performance on the XVNLI dataset, the visual entailment task from the IGLUE benchmark, for all available languages (AR: Arabic, ES: Spanish, FR: French, RU: Russian) are reported in Table~\ref{tab:xvnli_results}. 
\clicotea outperforms other models by a significant margin for all languages, except Russian where CCLM\textsubscript{3M} achieves comparable performance.

\begin{table}[h!]
\ra{1.3}
\centering
\begin{tabular}{@{}lrrrr@{}}
\toprule
 \textbf{Model} & \multicolumn{4}{c}{\textbf{Language}}\\
\cmidrule{2-5}
 & \multicolumn{1}{c}{AR} & \multicolumn{1}{c}{ES} & \multicolumn{1}{c}{FR} & \multicolumn{1}{c}{RU}  \\\midrule
mUNITER & 46.73 &56.96 & 59.36&  51.72\\
xUNITER & 51.98& 58.94 &63.32 & 59.71\\
UC${^2}$ & 56.19& 57.47&  69.67 & 64.86\\
M${^3}$P& 55.24& 58.85 &56.36 & 62.54\\
CCLM\textsubscript{3M} & 71.04& 75.80& 78.14& \textbf{73.56} \\
CCLM\textsubscript{4M} & 69.68 & 73.65 & 77.54 &72.40 \\
\midrule
\clicotea & \textbf{75.83} & \textbf{83.48 } & \textbf{80.17} & 73.13 \\
% \midrule
% Translation test & \multicolumn{5}{c}{93.73}\\
\bottomrule
\end{tabular}
\caption{Zero-shot performance on visual entailment with XVNLI dataset. Accuracy is reported.}
\label{tab:xvnli_results}
\end{table}

\subsection{In-domain vs Open-domain Data}

\begin{table}[h]
    \centering
    \begin{tabular}{@{}lrr@{}}
    \toprule
        \multicolumn{1}{c}{Language} & \multicolumn{1}{c}{Total number }& \multicolumn{1}{c}{Accuracy} \\ 
        &\multicolumn{1}{c}{of sentences}& \multicolumn{1}{c}{in \%}\\
        \midrule
        Swahili & 50400  & 63.27 \\
        Turkish & 50418 & 66.61\\
        Chinese & 51159 & 59.09\\
        \bottomrule
    \end{tabular}
    \caption{Zero-shot performance on visual reasoning with MaRVL dataset. Alignment is done with a subset from XNLI dataset.}
    \label{tab:xnli_results}
\end{table}

In order to eliminate the need for machine translations from \clicotea in step~\ref{pipeline:corpus}, we created a parallel text corpus with sentences obtained from XNLI~\cite{conneau2018xnli} which is publicly available and covers 15 languages. A subset of XNLI has been used for training the alignment by considering only the sentences that were semantically close to the captions in NLVR2. To do so, we used the \texttt{Sentence-Transformers} framework\footnote{Available at \url{ https://www.sbert.net}.} to compute sentence embeddings similarities between NLVR2 captions and XNLI English sentences and kept only the sentences with a cosine similarity higher than 0.5. About 50k English sentences from XNLI are semantically close to NLVR2 captions, we thus selected their parallel sentences in Swahili, Turkish and Chinese to perform an evaluation on MaRVL dataset.  
After the contextualised token alignment training on XNLI-based datasets,  our results in Table~\ref{tab:xnli_results} suggest that a multilingual open-domain dataset gives better results than mUNITER and xUNITER but underperforms the results obtained by translating in-domain training sets. This could be explained by the fact that although these datasets are multilingual, the sentences are not semantically close enough to NLVR2 captions.

 %Hence, in order to eliminate machine translation engines that could give bad translations and contribute to accumulate error, an in-domain dataset similar to NLVR2 should be created. This dataset should not only include complex relationships that are beneficial for reasoning tasks, but also references to images.

%% file: acl_latex.bbl
\begin{thebibliography}{21}
\expandafter\ifx\csname natexlab\endcsname\relax\def\natexlab#1{#1}\fi

\bibitem[{Bugliarello et~al.(2022)Bugliarello, Liu, Pfeiffer, Reddy, Elliott,
  Ponti, and Vuli{\'c}}]{bugliarello2022iglue}
Emanuele Bugliarello, Fangyu Liu, Jonas Pfeiffer, Siva Reddy, Desmond Elliott,
  Edoardo~Maria Ponti, and Ivan Vuli{\'c}. 2022.
\newblock Iglue: A benchmark for transfer learning across modalities, tasks,
  and languages.
\newblock \emph{arXiv preprint arXiv:2201.11732}.

\bibitem[{Chen et~al.(2020)Chen, Li, Yu, El~Kholy, Ahmed, Gan, Cheng, and
  Liu}]{chen2020uniter}
Yen-Chun Chen, Linjie Li, Licheng Yu, Ahmed El~Kholy, Faisal Ahmed, Zhe Gan,
  Yu~Cheng, and Jingjing Liu. 2020.
\newblock Uniter: Universal image-text representation learning.
\newblock In \emph{European conference on computer vision}, pages 104--120.
  Springer.

\bibitem[{Conneau et~al.(2019)Conneau, Khandelwal, Goyal, Chaudhary, Wenzek,
  Guzm{\'a}n, Grave, Ott, Zettlemoyer, and Stoyanov}]{conneau2019unsupervised}
Alexis Conneau, Kartikay Khandelwal, Naman Goyal, Vishrav Chaudhary, Guillaume
  Wenzek, Francisco Guzm{\'a}n, Edouard Grave, Myle Ott, Luke Zettlemoyer, and
  Veselin Stoyanov. 2019.
\newblock Unsupervised cross-lingual representation learning at scale.
\newblock \emph{arXiv preprint arXiv:1911.02116}.

\bibitem[{Conneau et~al.(2018)Conneau, Rinott, Lample, Williams, Bowman,
  Schwenk, and Stoyanov}]{conneau2018xnli}
Alexis Conneau, Ruty Rinott, Guillaume Lample, Adina Williams, Samuel~R.
  Bowman, Holger Schwenk, and Veselin Stoyanov. 2018.
\newblock Xnli: Evaluating cross-lingual sentence representations.
\newblock In \emph{Proceedings of the 2018 Conference on Empirical Methods in
  Natural Language Processing}. Association for Computational Linguistics.

\bibitem[{Devlin et~al.(2018)Devlin, Chang, Lee, and
  Toutanova}]{devlin2018bert}
Jacob Devlin, Ming-Wei Chang, Kenton Lee, and Kristina Toutanova. 2018.
\newblock Bert: Pre-training of deep bidirectional transformers for language
  understanding.
\newblock \emph{arXiv preprint arXiv:1810.04805}.

\bibitem[{Dou and Neubig(2021)}]{dou2021word}
Zi-Yi Dou and Graham Neubig. 2021.
\newblock Word alignment by fine-tuning embeddings on parallel corpora.
\newblock \emph{arXiv preprint arXiv:2101.08231}.

\bibitem[{Hu et~al.(2020)Hu, Ruder, Siddhant, Neubig, Firat, and
  Johnson}]{hu2020xtreme}
Junjie Hu, Sebastian Ruder, Aditya Siddhant, Graham Neubig, Orhan Firat, and
  Melvin Johnson. 2020.
\newblock Xtreme: A massively multilingual multi-task benchmark for evaluating
  cross-lingual generalisation.
\newblock In \emph{International Conference on Machine Learning}, pages
  4411--4421. PMLR.

\bibitem[{Li et~al.(2022)Li, Li, Xiong, and Hoi}]{li2022blip}
Junnan Li, Dongxu Li, Caiming Xiong, and Steven Hoi. 2022.
\newblock Blip: Bootstrapping language-image pre-training for unified
  vision-language understanding and generation.
\newblock \emph{arXiv preprint arXiv:2201.12086}.

\bibitem[{Li et~al.(2021)Li, Selvaraju, Gotmare, Joty, Xiong, and
  Hoi}]{li2021align}
Junnan Li, Ramprasaath Selvaraju, Akhilesh Gotmare, Shafiq Joty, Caiming Xiong,
  and Steven Chu~Hong Hoi. 2021.
\newblock Align before fuse: Vision and language representation learning with
  momentum distillation.
\newblock \emph{Advances in neural information processing systems},
  34:9694--9705.

\bibitem[{Lin et~al.(2014)Lin, Maire, Belongie, Hays, Perona, Ramanan,
  Doll{\'a}r, and Zitnick}]{lin2014microsoft}
Tsung-Yi Lin, Michael Maire, Serge Belongie, James Hays, Pietro Perona, Deva
  Ramanan, Piotr Doll{\'a}r, and C~Lawrence Zitnick. 2014.
\newblock Microsoft coco: Common objects in context.
\newblock In \emph{European conference on computer vision}, pages 740--755.
  Springer.

\bibitem[{Ni et~al.(2021)Ni, Huang, Su, Cui, Bharti, Wang, Zhang, and
  Duan}]{ni2021m3p}
Minheng Ni, Haoyang Huang, Lin Su, Edward Cui, Taroon Bharti, Lijuan Wang,
  Dongdong Zhang, and Nan Duan. 2021.
\newblock M3p: Learning universal representations via multitask multilingual
  multimodal pre-training.
\newblock In \emph{Proceedings of the IEEE/CVF conference on computer vision
  and pattern recognition}, pages 3977--3986.

\bibitem[{Pfeiffer et~al.(2021)Pfeiffer, Geigle, Kamath, Steitz, Roth,
  Vuli{\'c}, and Gurevych}]{pfeiffer2021xgqa}
Jonas Pfeiffer, Gregor Geigle, Aishwarya Kamath, Jan-Martin~O Steitz, Stefan
  Roth, Ivan Vuli{\'c}, and Iryna Gurevych. 2021.
\newblock xgqa: Cross-lingual visual question answering.
\newblock \emph{arXiv preprint arXiv:2109.06082}.

\bibitem[{Pires et~al.(2019)Pires, Schlinger, and
  Garrette}]{pires2019multilingual}
Telmo Pires, Eva Schlinger, and Dan Garrette. 2019.
\newblock How multilingual is multilingual bert?
\newblock \emph{arXiv preprint arXiv:1906.01502}.

\bibitem[{Plummer et~al.(2015)Plummer, Wang, Cervantes, Caicedo, Hockenmaier,
  and Lazebnik}]{plummer2015flickr30k}
Bryan~A Plummer, Liwei Wang, Chris~M Cervantes, Juan~C Caicedo, Julia
  Hockenmaier, and Svetlana Lazebnik. 2015.
\newblock Flickr30k entities: Collecting region-to-phrase correspondences for
  richer image-to-sentence models.
\newblock In \emph{Proceedings of the IEEE international conference on computer
  vision}, pages 2641--2649.

\bibitem[{Schwenk et~al.(2021)Schwenk, Chaudhary, Sun, Gong, and
  Guzm{\'a}n}]{schwenk-etal-2021-wikimatrix}
Holger Schwenk, Vishrav Chaudhary, Shuo Sun, Hongyu Gong, and Francisco
  Guzm{\'a}n. 2021.
\newblock \href {https://doi.org/10.18653/v1/2021.eacl-main.115}
  {{W}iki{M}atrix: Mining 135{M} parallel sentences in 1620 language pairs from
  {W}ikipedia}.
\newblock In \emph{Proceedings of the 16th Conference of the European Chapter
  of the Association for Computational Linguistics: Main Volume}, pages
  1351--1361, Online. Association for Computational Linguistics.

\bibitem[{Suhr et~al.(2017)Suhr, Lewis, Yeh, and Artzi}]{suhr2017corpus}
Alane Suhr, Mike Lewis, James Yeh, and Yoav Artzi. 2017.
\newblock A corpus of natural language for visual reasoning.
\newblock In \emph{Proceedings of the 55th Annual Meeting of the Association
  for Computational Linguistics (Volume 2: Short Papers)}, pages 217--223.

\bibitem[{Wu and Dredze(2019)}]{wu2019beto}
Shijie Wu and Mark Dredze. 2019.
\newblock Beto, bentz, becas: The surprising cross-lingual effectiveness of
  bert.
\newblock \emph{arXiv preprint arXiv:1904.09077}.

\bibitem[{Yasunaga et~al.(2022)Yasunaga, Bosselut, Ren, Zhang, Manning, Liang,
  and Leskovec}]{yasunaga2022deep}
Michihiro Yasunaga, Antoine Bosselut, Hongyu Ren, Xikun Zhang, Christopher~D
  Manning, Percy Liang, and Jure Leskovec. 2022.
\newblock Deep bidirectional language-knowledge graph pretraining.
\newblock \emph{arXiv preprint arXiv:2210.09338}.

\bibitem[{Zeng et~al.(2022)Zeng, Zhou, Luo, and Zhang}]{zeng2022cross}
Yan Zeng, Wangchunshu Zhou, Ao~Luo, and Xinsong Zhang. 2022.
\newblock Cross-view language modeling: Towards unified cross-lingual
  cross-modal pre-training.
\newblock \emph{arXiv preprint arXiv:2206.00621}.

\bibitem[{Zhang et~al.(2021)Zhang, Bosselut, Yasunaga, Ren, Liang, Manning, and
  Leskovec}]{zhang2021greaselm}
Xikun Zhang, Antoine Bosselut, Michihiro Yasunaga, Hongyu Ren, Percy Liang,
  Christopher~D Manning, and Jure Leskovec. 2021.
\newblock Greaselm: Graph reasoning enhanced language models.
\newblock In \emph{International Conference on Learning Representations}.

\bibitem[{Zhou et~al.(2021)Zhou, Zhou, Wang, Cheng, Li, Yu, and
  Liu}]{zhou2021uc2}
Mingyang Zhou, Luowei Zhou, Shuohang Wang, Yu~Cheng, Linjie Li, Zhou Yu, and
  Jingjing Liu. 2021.
\newblock Uc2: Universal cross-lingual cross-modal vision-and-language
  pre-training.
\newblock In \emph{Proceedings of the IEEE/CVF Conference on Computer Vision
  and Pattern Recognition}, pages 4155--4165.

\end{thebibliography}
